%% file: paper.tex
\documentclass[11pt,a4paper]{article}
\usepackage[hyperref]{eacl2021}
\usepackage{times}
\usepackage{latexsym}
\usepackage{amssymb}
\usepackage{url}
\usepackage{examples}
\usepackage{multirow}
\usepackage{color}
\usepackage{xcolor,colortbl}
\usepackage{graphicx}
\usepackage{amsmath}
\definecolor{lightgrey}{RGB}{227,226,219}

\usepackage{microtype}

\newcommand{\squishlist}{
 \begin{list}{$\bullet$}
  { \setlength{\itemsep}{0pt}
     \setlength{\parsep}{3pt}
     \setlength{\topsep}{3pt}
     \setlength{\partopsep}{0pt}
     \setlength{\leftmargin}{1.5em}
     \setlength{\labelwidth}{1em}
     \setlength{\labelsep}{0.5em} } }

\newcommand{\squishend}{
  \end{list}  }

\aclfinalcopy

\title{TDMSci:  
A Specialized Corpus for Scientific Literature Entity Tagging of Tasks Datasets and Metrics}

\author{Yufang Hou, Charles Jochim, Martin Gleize, Francesca Bonin and Debasis Ganguly \\
IBM Research Europe, Ireland \\
  \texttt{\{yhou|mgleize|fbonin|debasga1\}@ie.ibm.com} \\}

\date{}

\begin{document}
\maketitle

\begin{abstract}
\emph{Tasks}, \emph{Datasets} and \emph{Evaluation Metrics} are important concepts for understanding experimental scientific papers. However, most previous work on information extraction for scientific literature 
mainly focuses on the abstracts only, and does not treat datasets as a separate type of entity \cite{Zadeh2016TheAR,Luan2018}. 
In this paper, we present a new corpus that contains domain expert annotations for \emph{Task (T), Dataset (D), Metric (M)} entities on 2,000 sentences extracted from NLP papers. We report experiment results on TDM extraction using a simple data augmentation strategy and apply our tagger to around 30,000 NLP papers from the ACL Anthology. 
The corpus is made publicly available to the community for fostering research on scientific publication summarization \cite{erera-etal-2019-summarization} 
and knowledge discovery.
\end{abstract}

 \input{intro}

\bibliographystyle{acl_natbib}
\bibliography{kbp}

\newpage
\appendix
\section{TDM Entity Annotation Guidelines}

\input{AnnotationGuidelines_embed}

\end{document}

%% file: intro.tex
\section{Introduction}
The past years have witnessed a significant growth in the number of scientific publications and  benchmarks in many disciplines. As an example, in the year 2019 alone, more than 170k papers were
submitted to the pre-print repository arXiv\footnote{\url{https://arxiv.org/help/stats/2019_by_area}}
 and among them, close to 10k papers were classified as NLP papers (i.e., \texttt{cs.CL}). Each experimental scientific field, including NLP, will benefit from the massive increase in studies, benchmarks, and evaluations, as they can provide ingredients for novel scientific advancements. 

However, researchers may struggle to keep track of all studies published in a particular field, 
 resulting in duplication of research, comparisons with old or outdated benchmarks, and lack of progress. 
In order to tackle this problem, 
recently there have been a few manual efforts to summarize the state-of-the-art on selected subfields of NLP in the form of leaderboards that extract tasks, datasets, metrics and results from papers, such as \emph{NLP-progress}\footnote{\url{https://github.com/sebastianruder/NLP-progress}} or \emph{paperswithcode}.\footnote{\url{https://paperswithcode.com}}
But these manual efforts are not sustainable over time for all NLP tasks. 

Over the past few years, several studies and shared tasks have begun to tackle the task of entity extraction from scientific papers.
\newcite{semeval2017} formalized a task 
to identify three types of entities (i.e., \emph{task, process, material}) in scientific publications (SemEval 2017 task10). 
 \newcite{semeval2018} presented a task (SemEval 2018 task 7) on semantic relation extraction from NLP papers. They provided a dataset of 350 abstracts and reuse the entity annotations from \newcite{Zadeh2016TheAR}. Recently \newcite{Luan2018} released a corpus containing 500 abstracts with six types of entity annotations. However, these corpora do not treat \emph{Dataset} as a separate type of entity and most of them focus on the abstracts only.

In a previous study, we developed an IE system to extract \{\emph{task, dataset, metric}\} triples from NLP papers based on a small, manually created 
task/dataset/metric (TDM) taxonomy \cite{Hou-acl2019}. In practice, we found that a TDM
knowledge base is required to extract TDM information and build NLP leaderboards for a wide range of NLP papers. This can help researchers quickly understand related literature for a particular task, or to perform comparable experiments.

As a first step to build such a TDM knowledge base for the NLP domain, in this paper we present a specialized English corpus containing 
2,000 sentences taken from the full text of NLP papers which have been annotated by domain experts for three main concepts: \textit{Task} (T), \textit{Dataset} (D) and \textit{Metric} (M). Based on this corpus, we develop a TDM tagger using a novel data augmentation technique. In addition, we apply this tagger to around 30,000 NLP papers from the ACL Anthology and demonstrate its value to construct an NLP TDM knowledge graph.
We release our corpus at \url{https://github.com/IBM/science-result-extractor}.

 \section{Related Work}
  A lot of interest has been focused on information extraction from scientific literature. 
SemEval 2017-task 10 \cite{semeval2017} 
proposed a new  task for the identification of three types of entities (\textit{Task, Method}, and \textit{Material}) 
in a corpus of 
500 paragraphs taken from open access journals. 
Based on \newcite{semeval2017} and \newcite{semeval2018}, \newcite{Luan2018} created \emph{SciERC}, a dataset containing 500 scientific abstracts with annotations for six types of entities and relations between them. Both SemEval 2017-task 10 and \emph{SciERC} do not treat ``\emph{dataset}'' as a separate entity type. Instead, their ``\emph{material}'' category comprises a much larger set of resource types, including tools, knowledge resources, bilingual dictionaries, as well as datasets. In our work, we focus on ``\emph{datasets}” entities that researchers use to evaluate their approaches because dataset is one of the three core elements to construct leaderboards for NLP papers. 

Concurrent to our work, \newcite{jain-etal-2020-scirex} develop a new corpus \emph{SciREX} which contains 438 papers on different domains from \emph{paperswithcode}. It includes annotations for four types of entities (i.e., \emph{Task, Dataset, Metric, Method}) and the relations between them. The initial annotations were carried out automatically using distant signals from \emph{paperswithcode}. Later human annotators performed necessary
corrections to generate the final dataset. \emph{SciREX} is the closest to our corpus in terms of entity annotations. In our work, we focus on TDM entities which reflect the collectively shared views in the NLP community and 
our corpus is annotated by five experts who all have 5-10 years NLP research experiences.

\section{Corpus Creation}
 \label{sec:corpus_creation}

 \subsection{Annotation Scheme}
 We developed an annotation scheme for annotating Task, Dataset, and Evaluation Metric phrases in NLP papers. Our annotation guidelines\footnote{Please see the appendix for the whole annotation scheme.} are based on the scientific term annotation scheme described in \newcite{Zadeh2016TheAR}. Different from previous corpora \cite{Zadeh2016TheAR,Luan2018}, we only annotated \textbf{factual} and \textbf{content-bearing} entities. This is because we aim to build a TDM knowledge base in the future and 
 non-factual entities (e.g., \emph{a high-coverage sense-annotated corpus} in Example \ref{exa:exa1}) do not reflect the collectively shared views of TDM entities in the NLP domain.
\begin{examples}
\item \label{exa:exa1} In order to learn models for disambiguating a large set of content words, \emph{a high-coverage sense-annotated corpus} is required.
\end{examples}  
 
Following the above guidelines, we also do not annotate \emph{anonymous entities}, such as ``\emph{this task}'' or ``\emph{the dataset}''. These entities are anaphors and can not be used independently to refer to any specific TDM entities without contexts. 
In general, we choose to annotate TDM entities that normally have specific names and whose meanings usually are consistent across different papers. From this perspective, the TDM entities that we annotate are similar to named entities, which are self-sufficient to identify the referents.

  \subsection{Pilot Annotation Study}
\paragraph{Data preparation.}
For the pilot annotation study, we choose 100 sentences from the NLP-TDMS corpus \cite{Hou-acl2019}. The corpus contains 332 NLP papers which are annotated with triples of \emph{\{Task, Dataset, Metric\}} on the document level. We use string and substring match to extract a list of sentences from these papers which are likely to contain the document level \emph{Task, Dataset, Metric} annotations. We then manually choose 100 sentences from this list following the criteria: 1) the sentence should contain the valid mention of \emph{Task}, \emph{Dataset}, or \emph{Metric}; 2) the sentences should come from different papers as much as possible; and 3) there should be a balanced distribution of \emph{task}, \emph{dataset}, and \emph{metric} mentions in these sentences.

\paragraph{Annotation agreement.} Four NLP domain experts annotated the same 100 sentences for a pilot annotation study, following the annotation guidelines described above. All the annotations were conducted using BRAT \cite{stenetorp-etal-2012-brat}. 
The inter annotator agreement has been calculated with a pairwise comparison between annotators using \emph{precision}, \emph{recall} and \emph{F-score} on the exact match of the annotated entities. 
In other words, two entities are considered matching (true positive) if they have the same boundaries and are assigned to the same label. We also calculate Fleiss' kappa on a per token basis, comparing the agreement of annotators on each token in the corpus. Table \ref{tab:agreement} lists the mean F-score as well as the token-based Fleiss' $\kappa$ value for each entity type.
Overall, we achieve high reliability for all categories.

\begin{table}[t]
\begin{center}
\begin{tabular}{l|c|c}
&Mean F-score &Fleiss' $\kappa$  \\ 
&(EM)& (Token) \\ \hline
Task   &0.720&0.797\\ 
Dataset   &0.752&0.829\\ 
Metric   &0.757&0.896\\ \hline
Overall & 0.743&0.842
\end{tabular}
\end{center}
\caption{\label{tab:agreement}  Inter-annotator agreement.}
\end{table}

\paragraph{Adjudication.}
The final step of the pilot annotation was to reconcile 
 disagreements among the four annotators 
to produce the final canonical annotation. 
This step also allows us to refine the annotation guidelines. Specifically,  through the discussion of annotation disagreements we could identify ambiguities and omissions in the guidelines.
For example, one point of ambiguity was whether a \textit{task} must be associated with a dataset, or can we annotate higher level tasks, e.g., \textit{sequence labeling}, which do not have a dedicated dataset but may include several tasks and datasets.
This discussion also revealed the overlap in how we refer to tasks and datasets in the literature.
As authors we frequently use these interchangeably, often with shared tasks, e.g., ``\emph{SemEval-07 task 17}'' seems to more often refer to a dataset than a specific instance of the (Multilingual) Word Sense Disambiguation task, or the ``\emph{MultiNLI}'' corpus is sometimes used as shorthand for the task. 
After the discussion, we agreed that we should annotate higher level tasks. In addition, we should assign labels to entities according to their actual referential meanings in contexts.

\subsection{Main Annotation}
After the pilot study, 1,900 additional sentences were annotated by five NLP researchers.
Four annotators participated in the pilot annotation study, and all annotators joined  the adjudication discussion. 
Note that every annotator annotate different set of sentences. The annotator who designed the annotation scheme annotated 700 sentences, the other four annotators annotated 300 sentences each.\footnote{Due to time constraints, we did not carry out another round of pilot study. Partially it is because we felt that the revised guidelines resulting from the discussion were sufficient for the annotators to decide ambiguous cases. So in the second stage annotators annotated disjoint sets of sentences. After this, the annotator who designed the annotation scheme went through the whole corpus again to verify the annotations.}

In general, most sentences in our corpus are not from the abstracts. Note that the goal of developing our corpus is to automatically build an NLP TDM  taxonomy and use them to tag NLP papers. Therefore, the inclusion of sentences from the whole paper other than the abstract section is important for our purpose. Because not all abstracts talk about all three elements. For instances, for the top ten papers listed in the \{\emph{sentiment analysis, IMDB, accuracy}\} leaderboard in \emph{paperswithcode}\footnote{https://paperswithcode.com/sota/sentiment-analysis-on-imdb, search was carried out on November, 2020.}, only four abstracts mention the dataset ``\emph{IMDB}''. If we only focus on the abstracts, we will miss the other six papers from the leaderboard. 

\begin{table}
\begin{center}
\begin{tabular}{lcc}
 
&Train&Test\\ \hline
\# Sentences & 1500&500\\
\hline
\# Task   &1219&396\\ 
\# Dataset   &420&192\\ 
\# Metric   &536&174\\ 
\hline
\end{tabular}
\end{center}
\caption{\label{tab:expdata}  Statistics of task/dataset/metric mentions in the training and testing datasets.}
\end{table}

\begin{table*}[t]
\begin{center}
\begin{footnotesize}
\begin{tabular}{|l|c|c|c|c|c|c|c|c|c|c|c|c|}
\hline
 &  \multicolumn{3}{c|}{\emph{CRF}} &\multicolumn{3}{c|}{\emph{CRF w/ gazetteer}} & \multicolumn{3}{c|}{\emph{SciIE}} & \multicolumn{3}{c|}{\emph{Flair-TDM}}\\
 & P & R & F & P & R & F& P & R & F& P & R & F\\ \hline

\multicolumn{13}{|c|}{\emph{Original training data}}\\ \hline
 {Task}    & 63.79 & 46.72 & 53.94 & 61.86 & 45.45 & 52.40 & 69.23&54.55&61.02&61.54&	54.55&	57.83\\
 {Dataset} & 65.42 & 36.46 & 46.82 & 65.45 & 37.50 & 47.68 & 66.97&38.02&48.50&52.66&	46.35&49.30\\
 {Metric}  & 80.00 & 66.67 & 72.73 & 80.95 & 68.39 & 74.14 & 77.99&71.26& 74.47 &76.33	&74.14&	75.22\\ \hline
 {Micro-}  & 68.45 & 48.69 & 56.90 & 67.70 & 48.69 & 56.64 & 71.21&54.20& 
61.55 &62.99	&56.96&	59.79\\ \hline
\multicolumn{13}{|c|}{\emph{Original Training data + Augmented masked training data}}\\ \hline
 {Task}    & 63.24	&43.43&	51.50 & 62.96	&42.93&	51.05 & 68.63&55.81&\textbf{61.56}&65.14	&53.79&	58.92\\
 {Dataset} & 62.38	&32.81&	43.00 & 64.71&	34.38&	44.90 & 55.43&50.52&52.86&59.15	&50.52&	\textbf{54.50}\\
 {Metric}  & 80.15&	62.64&	70.32 & 79.29&	63.79&	70.70 & 76.83&72.41& 74.56 &79.63	&74.14&	\textbf{76.79}\\ \hline
 {Micro-}  & 67.58 & 45.14 & 54.13 & 67.77 & 45.54 & 54.47 & 67.17&58.27& 
\textbf{62.40} &67.23	&57.61&	62.05\\ \hline
 
\hline
\end{tabular}
\end{footnotesize}
\end{center}
\caption{Results of different models for \emph{task/dataset/metric} entity recognition on \emph{TDMSci} test dataset.}
\label{tab:results}
\end{table*}

\section{A TDM Entity Tagger}
Our final corpus \emph{TDMSci} contains 2,000 sentences with 2,937 mentions of three entity types. We convert the original BRAT annotations to the standard CoNLL format using BIO scheme.\footnote{Note that our BRAT annotation contains a small amount of embedded entities, e.g., \emph{WSJ portion of Ontonotes} and \emph{Ontonotes}. We only keep the longest span when we convert the BRAT annotations to the CoNLL format.}
We develop a tagger to extract TDM entities based on this corpus.

\subsection{Experimental Setup}
To evaluate the performance of our tagger, we split \emph{TDMSci} into training and testing sets, which 
contains 1,500 and 500 sentences,  respectively. 
Table \ref{tab:expdata} shows the statistics of task/dataset/metric mentions in these two datasets. 
For evaluation, we report precision, recall, F-score on exact match for each entity type as well as micro-averaged precision, recall, F-score for all entities.

\subsection{Models}
\label{sec:model}

We model the task as a sequence tagging problem. We apply a traditional 
CRF model \cite{lafferty:etal:01} with various lexical features 
and a BiLSTM-CRF model for this task. To compare with the state-of-the-art entity extraction model on scientific literature, we also use \emph{SciIE} from \newcite{Luan2018} to train a \emph{TDM} entity recognition model based on our training data.  Below we describe all models in detail.

\paragraph{CRF.} 
We use the Stanford CRF implementation \cite{finkel-etal-2005-incorporating} to train a \emph{TDM} NER tagger based on our training data. 
We use the following features: 
unigrams of the previous, current and next words, current word character n-grams, current POS tag, surrounding POS tag sequence, current word shape, surrounding word shape sequence. 

\paragraph{CRF with gazetteers.}
To test whether the above CRF model can benefit from knowledge resources, 
we add two gazetteers to the feature set: one is a list containing around 6,000 dataset names which were crawled from LRE Map,\footnote{\url{http://www.elra.info/en/catalogues/lre-map/}} 
 and another gazetteer comprises around 30 common evaluation metrics compiled by the authors.

\paragraph{SciIE.} \newcite{Luan2018} proposed a multi-task learning system to extract entities and relations from scientific articles. \emph{SciIE} is based on span representations using ELMo \cite{peters-etal-2018-deep} and here we adapt it for \emph{TDM} entity extraction. Note that if \emph{SciIE} predicts several embedded entities, we keep the one that has the highest confidence score. 
In practice we notice that this does not happen in our corpus.

\paragraph{Flair-TDM} For BiLSTM-CRF model, we use the recent \emph{Flair} framework \cite{akbik2018coling} based on the cased BERT-base embeddings \cite{BERT}. 
We train our \emph{Flair-TDM} model
with a learning rate of 0.1, a batch size of 32, a hidden size of 768, and the maximum epochs of 150. 

\subsection{Data Augmentation}
\label{sec:datasugmentation}
For TDM entity extraction, we expect that the surrounding context will play an important role. For instance, in the following sentence ``we show that for X on the Y, our model outperforms the prior state-of-the-art'', one can easily guess that X is a task entity while Y is a dataset entity. As a result, we propose a simple data augmentation strategy that generates the additional mask training data by replacing every token within an annotated TDM entity as \texttt{UNK}.

   \begin{figure*}[t]
        \centering
        \frame{\includegraphics[width=0.75\textwidth]{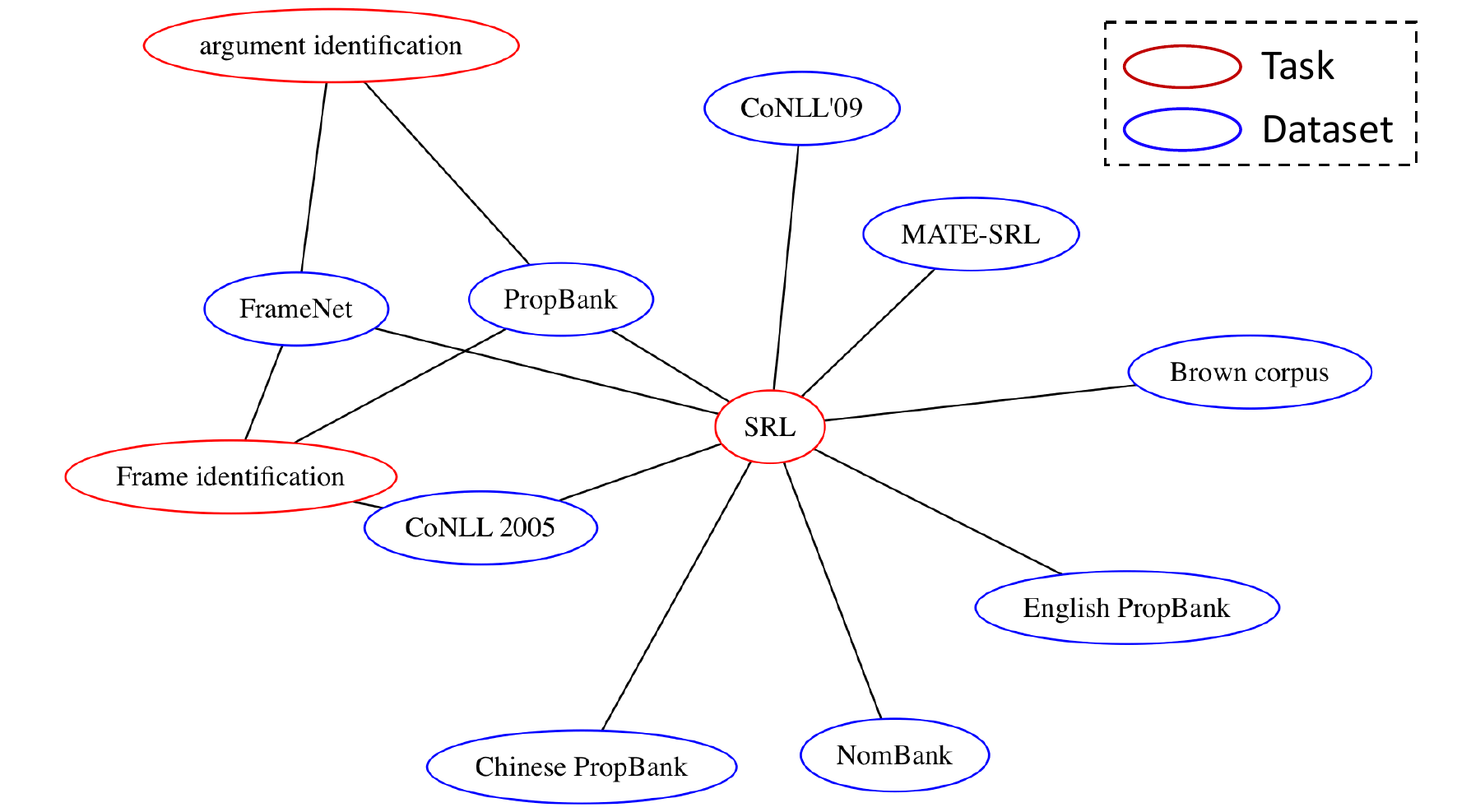}}
        \caption{A subset of the \emph{TDM} graph.}
        \label{fig:tdmgraph}
    \end{figure*}

\subsection{Results and Discussion}
Table \ref{tab:results} shows the performance of different models for \emph{task/dataset/metric} entity recognition on our testing dataset.

First, it seems that although adding gazetteers can help the CRF model detect \emph{dataset} and \emph{metric} entities better, the positive effect is limited. In general, both \emph{SciIE} and \emph{Flair-TDM} perform better than \emph{CRF} models for detecting all three type of entities.

Second, augmenting the original training data with the additional masked data as described in Section \ref{sec:datasugmentation} further improves the performance both for  \emph{SciIE} and \emph{Flair-TDM}. However, this is not the case for the CRF models. We assume this is because  CRF models heavily depend on the lexical features. 

Finally, we randomly sampled 100 sentences from the testing dataset and compared the predicted TDM entities in \emph{Flair-TDM} against the gold annotations. We found that most errors are from the boundary mismatch for task and dataset entities, e.g.,  \emph{text summarization} vs. \emph{abstractive text summarization}, or \emph{Penn Treebank} vs. \emph{Penn Treebank dataset}. The last error comes from the bias in the training data. A lot of researchers use ``\emph{Penn Treebank}'' to refer to a dataset. So the model will learn this bias and only tag "\emph{Penn Treebank}" as the dataset even though in a specific testing sentence, "\emph{Penn Treebank dataset}" was used to refer to the same corpus. 

In general, we think these mismatched predictions are reasonable in the sense that they capture the main semantics of the referents. Note that the numbers reported in Table \ref{tab:results} are based on exact match. Sometimes requiring exact match may be too restictive for downstreaming tasks. Therefore, we carried out an additional evaluation for the best \emph{Flair-TDM} model using partial match from SemEval 2013-Task 9 \cite{segura-bedmar-etal-2013-semeval}, which gives us a micro-average F1 of 76.47 for type partial match.

\section{An Initial TDM Knowledge Graph}
In this section, 
we apply the \emph{Flair-TDM} tagger to around 30,000 NLP papers from ACL Anthology to build an initial TDM knowledge graph. 

We downloaded all NLP papers from 1974 to 2019 that belong to ACL from the ACL Anthology\footnote{https://www.aclweb.org/anthology/}. For each paper, we collect sentences from the title, the abstract/introduction/dataset/corpus/experiment sections, as well as from the table captions. We then apply the \emph{Flair-TDM} tagger to these sentences. Based on the tagger results, we build an initial graph $G$ using the following steps:
\begin{itemize}
    \item add a \emph{TDM} entity as a node into $G$ if it appears at least five times in more than one paper;
    \item create a link between a \emph{task} node and a \emph{dataset/metric} node if they appear in the same sentence at least five times in different papers. 
\end{itemize}

By applying the above simple process, we get a noisy \emph{TDM} knowledge graph containing 180k nodes and 270k links. 
After checking a few dense areas, we find that our graph encodes valid knowledge about NLP task/dataset/metric. Figure \ref{fig:tdmgraph} shows that in our graph, the task ``SRL'' (semantic role labelling) is connected to a few datasets such as ``FrameNet'', ``PropBank'', and ``NomBank'' that are standard benchmark datasets for this task. 

Based on the tagged ACL Anthology and this initial noisy graph, we are exploring various methods to build a large-scale NLP TDM knowledge graph and to evaluate its accuracy/coverage in an ongoing work.

 \section{Conclusion}
\label{sec:conclusion}
In this paper, we have presented a new corpus (\emph{TDMSci}) annotated for three important concepts (\emph{Task/Dataset/Metric}) that are necessary for extracting the essential information from an NLP paper. Based on this corpus, we have developed a \emph{TDM} tagger using a simple but effective data augmentation strategy. Experiments on 30,000 NLP papers show that our corpus together with the \emph{TDM} tagger can help to build \emph{TDM} knowledge resources for the NLP domain.

%% file: AnnotationGuidelines_embed.tex
\subsection{Introduction}
This scheme describes guidelines for annotating \emph{Task}, \emph{Dataset}, and \emph{Evaluation Metric} phrases in NLP papers. We have pre-processed NLP papers in PDF format and chosen sentences that are likely to contain the above-mentioned entities for annotation. These sentences may come from different sections (e.g., Abstract, Introduction, Experiment, Dataset) as well as tables (e.g., table captions).

\subsection{Entity Types}
We annotate the following three entity types:
\begin{itemize}
    \item Task: A task is a problem to solve (e.g., \emph{information extraction}, \emph{sentiment classification}, \emph{dialog state tracking}, \emph{POS tagging}, \emph{NER}). 
    \item Dataset: A dataset is a specific corpus or language resource. Datasets are often used to develop models or run experiments for NLP tasks. A dataset normally has a short name, e.g., \emph{IMDB}, \emph{Gigaword}. 
    \item Metric: An evaluation metric explains the performance of a model for a specific task, e.g., \emph{BLEU} (for machine translation), or \emph{accuracy} (for a range of NLP tasks).  
\end{itemize}

\subsection{Notes and Examples}
\textbf{Entity spans.} Particular attention must be paid to the entity spans in order to improve agreement. The following list indicates all the annotation directions that annotators have been given regarding entity spans. Table~\ref{tab:spans} shows examples of correct span annotation.

\begin{itemize}
    \item Following the ACL RD-TEC 2.0 annotation guideline,\footnote{\url{https://github.com/languagerecipes/acl-rd-tec-2.0/blob/master/distribution/documents/acl-rd-tec-guidelines-ver2.pdf}} determiners should not be part of an entity span. For example, the string `the text8 test set`, only the span `test8' is annotated as \textit{dataset}. 
    \item 	Minimum span principle: Annotators should annotate only the minimum span necessary to represent the original meaning of task/dataset/metric. See Table \ref{tab:spans}, rows 1,2,3,4.
    \item Include `corpus/dataset/benchmark' when annotating dataset if these tokens are the head-noun of the dataset entity. For example: `ubuntu corpus', `SemEval-2010 Task 8 dataset'.
    \item 	Exclude the head noun of `task/problem' when annotating task (e.g., only annotation ``link prediction'' for ``the link prediction problem'') unless they are the essential part of the task itself (e.g., CoNLL-2012 shared task, SemEval-2010 relation classification task).
    \item Conjunction: If the conjunction NP is an ellipse, annotate the whole phrase (see Table~\ref{tab:spans}, rows 6,11); otherwise, annotate the conjuncts separately (see Table~\ref{tab:spans}, row 5).
    \item Tasks can be premodifiers (see Table~\ref{tab:spans}, rows 7,8,12)
    \item Embedded spans: Normally TDM entities do not contain any other TDM entities. A small number of \emph{Task} and \emph{Dataset} entities can contain other entities (see Table~\ref{tab:spans}, row 12). 
\end{itemize}

\begin{table*}[th]
    \centering
    \begin{tabular}{|c|p{6cm}|p{5cm}|l|} \hline
        Row & Phrase & Annotation & Entity \\
        \hline
    
        1 & The public Ubuntu Corpus & Ubuntu Corpus & Dataset \\
        \hline
        2 & the web portion of TriviaQA & web portion of TriviaQA & Dataset \\
        \hline
        3 & sentiment classification of movie reviews & sentiment classification & Task \\
        \hline
        4 & the problem of part-of-speech tagging  for informal, online conversational text & part-of-speech tagging & Task \\
        \hline
        5 & The FB15K and WN18 datasets & FB15K; WN18 & Dataset \\
        \hline
        6 & Hits at 1, 3 and 10 & Hits at 1, 3 and 10 & Metric \\
        \hline
        7 & Link prediction benchmarks & Link prediction & Task \\
        \hline
        8 & POS tagging accuracy & POS tagging; accuracy & Task, Metric \\
        \hline
        9 & the third Dialogue State Tracking Challenge & Dialogue State Tracking, third Dialogue State Tracking Challenge & Task, Dataset\\
        \hline
        10 & SemEval-2017 Task 9 & SemEval-2017 Task 9 & Task \\
        \hline
        11 & temporal and causal relation extraction and classification & 
          temporal and causal relation extraction and classification & Task \\
        \hline        
        12 & the SemEval-2010 Task 8 dataset & SemEval-2010 Task 8 dataset; SemEval-2010 Task 8 & Dataset,Task \\ \hline
    \end{tabular}
    \caption{Examples of entity span annotation guidelines}
    \label{tab:spans}
\end{table*}

\paragraph{Anonymous entities.} Do not annotate anonymous entities, which include anaphors. The following examples are anonymous entities:
\begin{itemize}
    \item \emph{this task}
    \item \emph{this metric}
    \item \emph{the dataset}
    \item \emph{a public corpus for context-sensitive response selection} in the sentence ``Experimental results in a a public corpus for context-sensitive response selection demonstrate the effectiveness of the proposed multi-vew model.''
\end{itemize}

\paragraph{Abbreviation.} If both the full name and the abbreviation are present in the sentence, annotate the abbreviation with its corresponding full name together. For instance, we annotate ``20-newsgroup (20NG)'' as a dataset entity in Example \ref{ex:ex1}.

\paragraph{Factual entity.} Only annotate ``factual, content-bearing'' entities. Task, dataset, and metric entities normally have specific names and their meanings are consistent across different papers. In Example \ref{ex:ex2}, ``\emph{a high-coverage sense-annotated corpus}'' is not a factual entity.

\begin{examples}
\item \label{ex:ex1} We used four datasets: IMDB, Elec, RCV1, and 20-newsgrous (20NG) to facilitate direct comparison with DL15.

\item \label{ex:ex2} In order to learn models for disambiguating a large set of content words, \underline{a high-coverage sense-annotated corpus} is required.

\end{examples}

%